\documentclass[10pt,twocolumn,letterpaper]{article}

\usepackage[pagenumbers]{cvpr}              %

\definecolor{cvprblue}{rgb}{0.21,0.49,0.74}
\usepackage[pagebackref,breaklinks,colorlinks,allcolors=cvprblue]{hyperref}

\usepackage{colortbl}
\usepackage{graphicx}
\usepackage{amsmath}
\usepackage{amssymb}
\usepackage{color}
\usepackage{booktabs}
\usepackage{enumitem}
\usepackage{tabularx} 
\usepackage{arydshln}
\usepackage{cuted}
\usepackage[table]{xcolor}
\definecolor{newlightblue}{RGB}{0,75,255}
\usepackage{array}
\usepackage[outline]{contour}
\usepackage{multirow}
\usepackage{xspace}
\usepackage[font=small]{caption}

\def\confName{CVPR}
\def\confYear{2026}

\newcommand{\trajvit}{TrajViT\xspace}
\newcommand{\modulename}{TrajTok\xspace}
\newcommand{\encodername}{TrajViT2\xspace}
\newcommand{\adaptername}{TrajAdapter\xspace}
\newcommand{\vlmname}{TrajVLM\xspace}

\definecolor{baselinecolor}{gray}{.9}
\newcommand{\baseline}[1]{\cellcolor{baselinecolor}{#1}}

\title{\modulename: Learning Trajectory Tokens Enhances Video Understanding}

\author{%
Chenhao Zheng$^{1,2}$, \
Jieyu Zhang$^{1,2}$, \
Jianing Zhang$^{1}$, 
Weikai Huang$^{1,2}$, 
Ashutosh Kumar$^{4}$, 
Quan Kong$^{4}$, \\
Oncel Tuzel$^{3}$, 
Chun-Liang Li$^{1,3}$, 
Ranjay Krishna$^{1,2}$
\\ 
\\
$^1$University of Washington,
$^2$Allen Institute for Artificial Intelligence,
$^3$Apple,
$^4$Woven by Toyota, Inc\\
\small{\url{https://github.com/hellomuffin/trajtok}}\\
 \vspace{-20pt}
}

\begin{document}

\twocolumn[{
\maketitle
\vspace{-10pt}
\begin{center}
\vspace{-10pt}
\includegraphics[width=0.95\textwidth]{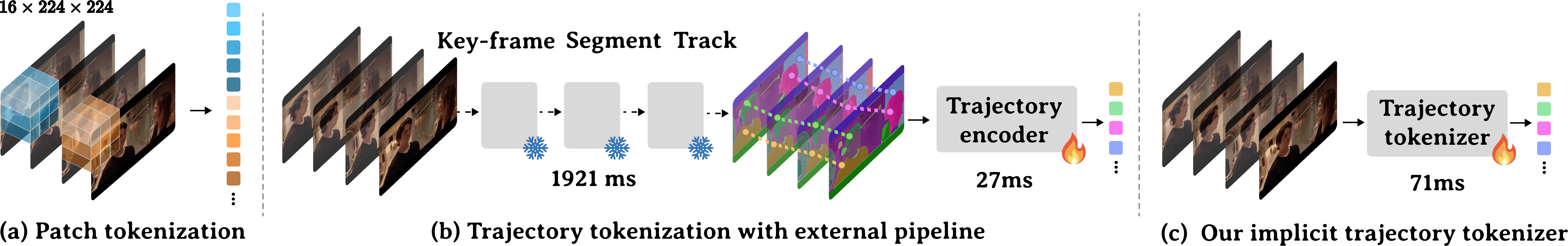}
\vspace{-5pt}
\captionof{figure}{(a) Traditional video tokenization splits a video into space-time patches, introducing large number of redundant tokens. (b) Prior work~\cite{zheng2025trajvit} proposes to represent a video via panoptic sub-object trajectory, which significantly reduces redundancy but relies on slow, non-differentiable pipelines.  (c) we propose \modulename{}, an end-to-end differentiable trajectory tokenizer that learns to implicitly propose trajectory tokens, offering low token counts, efficiency and adaptability to downstream objectives.}
\label{fig:teaser}
\end{center}
}]
\begin{abstract}
Tokenization in video models, typically through patchification, generates an excessive and redundant number of tokens. This severely limits video efficiency and scalability. 
While the recent trajectory-based tokenizers
offer a promising solution by decoupling video duration from token count, they rely on complex, external segmentation and tracking pipelines that are slow and task-agnostic. 
We propose \textbf{\modulename{}}, an end-to-end video tokenizer module that is fully integrated and co-trained with video models for a downstream objective, dynamically adapting its token granularity to semantic complexity, independent of video duration. 
\modulename{} contains a unified segmenter that performs implicit clustering over pixels in both space and time to directly produce object trajectories in a single forward pass. 
By prioritizing downstream adaptability over pixel-perfect segmentation fidelity,
\modulename{} is lightweight, efficient, and yet empirically improves video understanding performance.
With \modulename{}, we implement a video CLIP model trained from scratch (\textbf{\encodername{}}).  
It achieves the best accuracy at scale across both classification and retrieval benchmarks, 
while maintaining efficiency comparable to the best token-merging methods.
\modulename{} also proves to be a versatile component beyond its role as a tokenizer.  
We show that it can be seamlessly integrated as either a probing head for pretrained visual features (\textbf{\adaptername{}})
or an alignment connector in vision–language models (\textbf{\vlmname{}}) with especially strong performance in long-video reasoning.
\href{https://github.com/hellomuffin/trajtok}{Code is available at \texttt{github.com/hellomuffin/trajtok}}.

\end{abstract}
\section{Introduction}
\label{sec:intro}

Now that transformers are the dominant backbone in modern computer vision, designing effective tokenizers for visual inputs is a central research question~\cite{Beyer2023FlexiViT}. Tokenization for videos is particularly challenging due to their long duration and large number of near-duplicate frames. 
Today’s de-facto tokenization algorithms split the video tensor into space–time patches (Figure~\ref{fig:teaser}(a)). 
Whether training a ViT directly on raw video frames~\cite{Wang2023InternVid,vivit,timesformer}, adapting a pretrained vision encoder’s representations for downstream tasks~\cite{Bardes2024VJEPA,assran2025v}, or feeding visual tokens into a large vision–language model~\cite{Qwen2024Qwen25VL,Deitke2024Molmo}, visual tokens are almost invariably represented as regular grids of patches.
However, this fixed and spatially uniform tokenization becomes increasingly inefficient as the resolution or length of the video grows, leading to severe memory bottlenecks~\cite{vivit}.

Tokenization in video models, typically via simple patchification, produces an excessive number of spatio-temporal tokens, significantly limiting efficiency and scale. 
Recent token reduction efforts, which group semantically similar regions, often fail either by requiring predefined token counts~\cite{tome,Choudhury2024RLT,Mei2024SPFormer}, preventing adaptation to input complexity~\cite{tokenlearner}; or by compromising robustness due to sensitivity to scene motion~\cite{tome,Choudhury2024RLT,Choudhury2025APT}.
A more compelling alternative, \trajvit{}~\cite{zheng2025trajvit} introduced a promising paradigm by treating sub-object trajectories as the fundamental unit of video tokenization (Figure~\ref{fig:teaser}(b)). Trajectory-based tokenization effectively decouples video duration from the total token count, and, for the first time, demonstrates that tokens after grouping outperform raw patch tokens on all downstream tasks. However, this approach is fundamentally limited by its reliance on using external task-agnostic segmentation and tracking models~\cite{chen2024subobject,ravi2024sam2} to generate object trajectories, making the tokenizer a slow, independent, non-trainable preprocessing step.

We believe in the potential of organizing visual tokens according to object trajectories, as it closely aligns with human perceptual principles~\cite{Pylyshyn1988MOT,Spelke1990Principles,Wagemans2012Gestalt}; yet, we argue relying on an external pipeline to generate these trajectories is suboptimal.
Not only does it reduce efficiency and introduce longer latency, but it fixes the semantic granularity of the token unit using general-purpose segmentation models that may not be optimal for the downstream task. For instance, in understanding a particular dance performance, a model might require tokens representing dancers' individual body parts for fine-grained movements, whereas the task of identifying group formations might benefit from representing each dancer as a single, unified token. This mismatch motivates our goal: to build an implicit trajectory video tokenizer where the trajectory-generation module is seamlessly integrated into and co-trained with the rest of the network in an end-to-end manner, fully supervised by the downstream objective.

We present \textbf{\modulename{}}, an end-to-end video tokenizer that learns to group trajectories and proposes implicit trajectory tokens. 
\modulename{} is much more efficient than prior work~\cite{zheng2025trajvit} and is not rigid; 
it adapts its tokenization to the downstream tasks (Figure~\ref{fig:teaser}(c)).
Much of the compute in modern segmentation and tracking models is devoted to achieving pixel-perfect masks~\cite{cheng2022masked,XMem2022,Yang2023DeAOT,kirillov2023segment}, which is often superfluous for high-level understanding tasks.
By contrast, \modulename{} trades off pixel-perfect accuracy with end-task performance.
This is achieved by formulating trajectory generation as an implicit clustering problem over input pixels—both in space and in time.
By treating spatial and temporal dimensions uniformly, we design a unified segmenter that processes an entire video in one forward pass to directly output clusters of object trajectories.
Empirically, we show that reduced segmentation accuracy doesn't harm and instead improves understanding performance. 
Finally, our trajectory encoder incorporates adaptive representation inspired by Matryoshka~\cite{kusupati2022matryoshka}, enabling adaptive token number per trajectory and resolving the issue of over-compressed representations for objects undergoing complex or articulated motion.

\begin{figure*}[htbp]
    \centering
    \includegraphics[width=\linewidth]{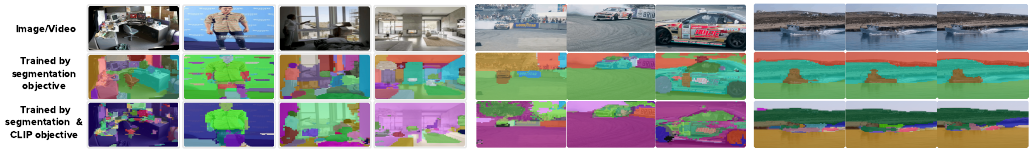}
    \caption{\textbf{Training with downstream understanding tasks reshapes the segmentation granularity.}  We visualize the trajectory masks produced by our segmenter when trained with only segmentation supervision versus jointly with segmentation and CLIP objectives.
The CLIP objective reshapes the segmentation granularity, producing finer foreground object masks while merging background regions.
}
    \label{fig:seg_clip}
\end{figure*}

With \modulename{}, we train \textbf{\encodername{}}, a transformer encoder from scratch using the CLIP objective~\cite{Radford2021CLIP,zheng2023exif,zheng2024iterated}, and evaluate it on classification and retrieval benchmarks.
Our approach achieves the best accuracy across both classification and retrieval benchmarks, including a large-margin improvement of +4.8\% on Kinetics-400 and +4.1\% on SSv2 over a standard video ViT, while being as efficient in inference FLOPs when compared against state-of-the-art token-merging methods~\cite{tome,Choudhury2024RLT,Choudhury2025APT}.
Furthermore, we observe better scaling trends than \trajvit{}~\cite{zheng2025trajvit} as training-dataset size increases, possibly because of our segmenter's flexibility in adapting to downstream tasks (Figure~\ref{fig:seg_clip}).

\modulename{} is versatile and more than just a tokenizer. 
We show that a pretrained \modulename{} can also be used in two other scenarios.
First, we design \textbf{\adaptername{}} as a feature adaptor, inserted after a pretrained ViT. We show \adaptername{} provides a  cost-effective way to enhance the probing performance of pretrained encoders in common video classification benchmarks without full fine-tuning.
Second, we design \textbf{\vlmname{}}, a vision-language model with \modulename{} as the alignment module. When positioned between a ViT and an LLM, \modulename{} improves video question-answering performance, especially for long-video questioning benchmarks.
Together, these results highlight the potential of our end-to-end trajectory tokenizer as a unified, efficient, and semantically grounded tokenization module for dozens of video understanding tasks.

\section{Related work}
\begin{figure*}[tbp]
    \centering
    \includegraphics[width=\linewidth]{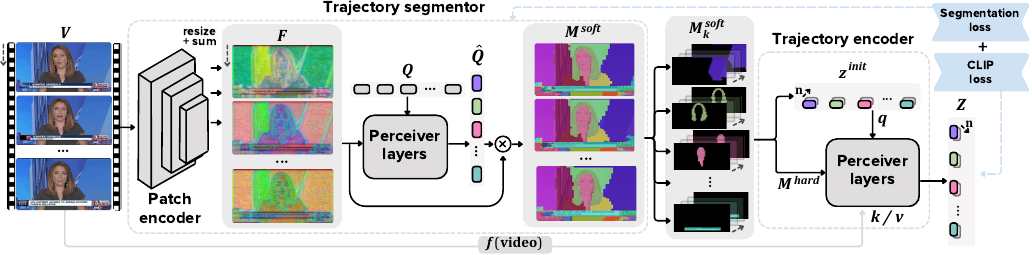}
   \caption{\textbf{Overview of the \modulename{} architecture.}  
\modulename{} comprises a \textit{trajectory segmenter} and a \textit{trajectory encoder}.  
The segmenter proposes trajectory masks for all objects in an image or video within a single forward pass.  
The encoder then aggregates raw video pixels or encoded visual features (parameterized by $f$ in the figure) according to these masks to produce trajectory tokens.  
The number of tokens per trajectory can be flexibly adjusted based on the available compute budget.
    }
    \label{fig:method}
\end{figure*}

\paragraph{Video tokenization and efficient video encoders.}  
Modern video transformers initially adopted fixed space–time patches~\cite{vivit,timesformer,liu2022videoswin}, but this leads to high token counts and heavy compute.  
To address this, a broad range of techniques has emerged, including token pruning and merging~\cite{rao2021dynamicvit,liang2022evit,fayyaz2022ats,tome,kim2024tokenfusion,wang2022stts,Choudhury2024RLT,choi2024vidtldr}, latent-bottleneck or learned-token approaches~\cite{tokenlearner,jaegle2021perceiver,jaegle2021perceiverio,nagrani2021bottlenecks}, and recent online or large-context video–LLM systems~\cite{wu2024videollmmod,prunevid,yan2025crosslmm,liu2025videoxlpro}.  
A persistent challenge across these efficiency-oriented designs is that their performance often lags behind patch-based tokenization, and scaling them to larger datasets or architectures remains difficult.  
More recently, trajectory-centric tokenization~\cite{zheng2025trajvit} has shown that organizing tokens by visual trajectories can simultaneously improve accuracy and reduce token counts.

\vspace{-10pt}
\paragraph{Object-centric representations.}  
Object-centric learning has long aimed to represent scenes as compositions of discrete entities rather than unstructured patches.
Early slot-based and scene decomposition models demonstrated the benefits of learning object-level structure from raw visual inputs \cite{locatello2020object,greff2019multiobject,burgess2019monet,kipf2021conditional}.
Foundation segmentation models such as SAM and SAM2 \cite{kirillov2023segment,ravi2024sam2} further enable region-level visual abstractions that improve grounding in vision–language models like Osprey \cite{li2024osprey}. In the context of video representation, both TrajViT~\cite{zheng2025trajvit} and Trokens \cite{kumar2025trokens} extend this object-centric perspective by introducing semantic, trajectory-based tokenization that groups spatio-temporal features into object-consistent units. 
Our work builds on this insight, generalizing trajectory-based tokenization into an end-to-end differentiable framework.

\section{\modulename{}}
We aim to design an end-to-end, efficient, and semantically grounded tokenizer that converts visual inputs (images or videos) into a compact set of tokens representing object trajectories. 
Let $\mathbf{V} \in \mathbb{R}^{T \times H \times W \times 3}$ denote an input video with $T$ frames and spatial resolution $H \times W$.
Our goal is to learn a mapping $\mathcal{T}: \mathbf{V} \rightarrow \mathbf{Z}$, where $\mathbf{Z} \in \mathbb{R}^{N \times d}$ is a set of $N$ trajectory tokens with dimension $d$. N is not fixed and depends on semantic complexity of the video.

The tokenizer consists of two differentiable components, which were trained jointly: a \textit{universal segmenter} that partitions the input into semantic groups, and a \textit{trajectory encoder} that aggregates these groups into compact latent tokens. We visualize the architecture in Figure~\ref{fig:method}.

\subsection{Universal segmenter for trajectory grouping}
The segmenter is a lightweight and efficient module that performs effective semantic grouping in a single feedforward pass. It decouples video duration from the final token count. We value robust semantic grouping over pixel-perfect segmentation masks, and design a simple and efficient module to achieve this.

\noindent\textbf{Frame-wise feature extraction.}
We first extract a high-resolution feature map from $\mathbf{V}$ using a lightweight patch encoder. We use ConvNeXt ~\cite{liu2022convnet} architecture as it naturally provides multi-scale feature maps.
We extract features frame-wise. Multi-scale maps are resized to their highest resolution ($1/4$ of original image size) and summed to form the final dense feature representation $\mathbf{F} \in \mathbb{R}^{T \times h \times w \times d}$, where $h, w = H/4, W/4$.

\noindent\textbf{Learnable queries for semantic grouping.}
We introduce a set of $N_q$ learnable latent queries $\mathbf{Q} \in \mathbb{R}^{N_q \times d}$ that act as cluster prototypes. These queries are processed through a stack of Perceiver~\cite{jaegle2021perceiver} layers. Within each perceiver layer, queries $\mathbf{Q}$ attend to the dense features $\mathbf{F}$ using cross-attention.
To handle inputs with variable frame counts $T$ and encode spatiotemporal structure, we apply 1D Rotary Positional Embeddings (RoPE)~\cite{su2024roformer} to the patch features $\mathbf{F}$ before attention.
The resulting processed queries, $\hat{\mathbf{Q}} = \text{Perceiver}(\mathbf{Q}, \text{RoPE}(\mathbf{F}))$, encapsulate the semantic information necessary for segmentation.

\noindent\textbf{Soft segmentation.}
We generate segmentation masks by computing the similarity between processed queries and patch features. A soft segmentation map $\mathbf{M}^{\text{soft}} \in [0, 1]^{N_q \times T \times h \times w}$ is obtained via softmax over the query dimension of the dot-product similarity:
\vspace{-5pt}
\begin{equation}
    \mathbf{M}^{\text{soft}}_{k, t, i, j} = \text{softmax}_k \left( \hat{\mathbf{q}}_k \cdot \mathbf{F}_{t,i,j} \right)
    \vspace{-5pt}
\end{equation}
where $\hat{\mathbf{q}}_k$ is the $k$-th processed query and $\mathbf{F}_{t,i,j}$ is the feature at time $t$ and spatial location $(i,j)$.
We find that feature maps at $1/4$ resolution provide sufficient detail for grouping, obviating the need for any compute-heavy decoders used in off-the-shelf segmenters~\cite{ye2025entitysam,cheng2022masked}.
Furthermore, we detach the gradients of $\mathbf{F}$ before entering the Perceiver layers to prevent unstable co-adaptation between patch features and learnable queries.

While the number of learnable queries $N_q$ is fixed, the number of trajectories $N$ can vary. 
Queries that produce empty masks are discarded, and long videos are divided into temporal chunks that can be processed in parallel.
This mechanism allows the tokenizer to propose a dynamic number of tokens that scales naturally with scene complexity.

\noindent\textbf{Training the segmenter.}
The segmenter can be trained either independently or jointly with the other objectives. 
We use supervised learning for the segmenter with pseudo ground-truth masks (generated via the TrajViT~\cite{zheng2025trajvit} pipeline). We find that a combination of \textit{Dice loss}~\cite{sudre2017generalised} and \textit{Focal loss}~\cite{lin2017focal}, without standard \textit{cross-entropy}, yields the best downstream understanding. This combination prioritizes the discovery of all object regions over strict pixel-level class accuracy.
This recipe likely arises because pixel-level precision is less critical for the downstream video benchmarks and applications we evaluate. 
In particular, Dice loss plays a central role in the discovery of all object regions within the visual input, ensuring robust semantic grouping.

\subsection{Trajectory encoder}
The trajectory encoder aggregates patch-level feature maps into compact tokens corresponding to segmented regions. 
Unlike prior approach~\cite{zheng2025trajvit}, our encoder accepts both soft and hard segmentations to ensure differentiability while maintaining disentangled representations. We default to use the patch enoder's features $F$ as the input feature map, but in practice it can be provided by any pretrained feature $f(video)$, enabling our tokenizer to operate as a plug-in feature adapter across diverse downstream tasks.

\noindent\textbf{Trajectory proposal generation.}
Initial trajectory embeddings are generated by a weighted aggregation of features using the soft masks. The proposal embedding $\mathbf{z}^{\text{init}}_k$ for the $k$-th trajectory is computed as:
\vspace{-5pt}
\begin{equation}
    \mathbf{z}^{\text{init}}_k = \sum_{t,i,j} \mathbf{M}^{\text{soft}}_{k,t,i,j} \cdot \mathbf{F}_{t,i,j}
    \vspace{-7pt}
\end{equation}
This soft aggregation allows gradients from downstream tasks to flow back into the segmenter. However, weighted summing can lead to information loss and blurred representations, which we address next.

\noindent\textbf{Trajectory embedding refinement.}
To sharpen these representations, we employ a second Perceiver module. The initial proposals $\mathbf{z}^{\text{init}}$ serve as cluster representations. We can now use them as queries to extract meaningful representations from $\mathbf{F}$.
To ensure disentanglement, we enforce \textit{masked} cross-attention using hard segmentation maps, $\mathbf{M}^{\text{hard}}$, obtained by applying an argmax to $\mathbf{M}^{\text{soft}}$ and converting to one-hot binary assignments.
The $k$-th query $\mathbf{z}^{\text{init}}_k$ is only allowed to attend to features $\mathbf{F}_{t,i,j}$ where $\mathbf{M}^{\text{hard}}_{k,t,i,j} = 1$. This refinement recovers fine-grained motion and texture details specific to the trajectory's  region.

\noindent\textbf{Adaptive token number per trajectory.}  
In practice, assigning a single token to each trajectory can be overly restrictive, especially for trajectories that span long durations, exhibit complex motion, or undergo substantial appearance changes.  
We introduce an adaptive token mechanism inspired by Matryoshka representations~\cite{kusupati2022matryoshka} to balance efficiency and expressivity.
Given a predefined compute budget, the encoder can emit $n \in \{1, 2, 4\}$ tokens per trajectory, enabling a flexible trade-off between efficiency and expressivity. 
This mechanism is applied during the Trajectory embedding refinement, where each trajectory token can be expanded into multiple sub-tokens.  We illustrate the details next.

For each of the initial trajectory embedding $z_k^\text{init}$, we duplicate the token $n$ times and associate each copy with a distinct learnable query vector.
During the subsequent attention step, these queries interact with the same set of patch features, allowing them to capture complementary aspects of the same trajectory.
However, we observe minimal performance gain with a naive initialization of these queries, as they tend to attend to the same dominant regions without explicit encouragement of diversity.  
To encourage diversity among sub-tokens, these queries are initialized with Fourier positional embeddings with angular offsets that maximally separate them in feature space.

To train such tokens, similar to Matryoshka Representations, we randomly sample $n\in\{1, 2, 4\}$ for each batch during training so that a single model can handle multiple token granularities.  
At inference, $n$ can be adjusted according to available computational resources.

\begin{figure}[tbp]
    \centering
    \includegraphics[width=\linewidth]{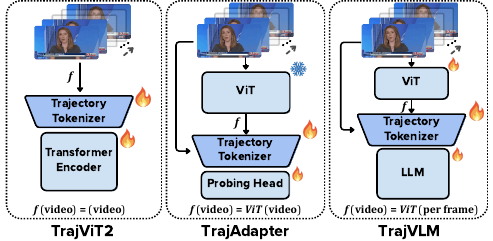}
   \caption{\textbf{\modulename{} is a versatile module applicable across pretraining, feature adaptation, and finetuning stages.}  
We demonstrate its use in three scenarios: \encodername{}, which trains a visual encoder from scratch; \adaptername{}, which adapts pretrained features for downstream tasks; and \vlmname{}, which uses \modulename{} as a connector in LLaVA-style large vision–language models.}
\vspace{-15pt}
    \label{fig:Applications}
\end{figure}

\section{Experiments}

In our experiments, we demonstrate that \textbf{\modulename{}} is a high-performance, efficient, and widely applicable module. It can operate directly on raw video pixels to propose trajectory tokens, or act as a feature adapter module applied to pretrained vision features.

We evaluate \modulename{} in three distinct scenarios, shown in Figure~\ref{fig:Applications}:
\begin{enumerate}
\item \textbf{\encodername{}}: a video transformer encoder trained from scratch under the CLIP objective, where the tokenizer directly proposes trajectory tokens from video pixels.
\item \textbf{\adaptername{}}: a plug-in feature adapter that aggregates dense feature maps from any pretrained video encoder, using trajectory-based grouping to yield more informative representations for downstream probing tasks.
\item \textbf{\vlmname{}}: a LLaVA-style~\cite{liu2023visual} video–language model in which  \modulename{} serves as a connector between a ViT and an LLM, grouping ViT features along trajectories and passing the grouped trajectory tokens as visual inputs to the language model.
\end{enumerate}

\noindent\textbf{{Tokenizer architecture}.} We use ConvNext-tiny~\cite{liu2022convnet} as the architecture for patch encoder. The perceiver modules in segmenter and trajectory encoder both have 2 layers and 8 attention heads. We use 128 learnable queries inside segmenter to cluster visual inputs into trajectories.  Ablations of these design choices are presented in Section~\ref{sec:abltion}.

\noindent\textbf{{Pretraining the segmenter}.}
When using ~\modulename{} in pretraining task where the training data is fully controllable, we initialize the segmenter from scratch and jointly train the tokenizer with other modules using an added segmentation loss.
However, when using the tokenizer as a feature adaptor in downstream tasks, we do not assume access to large-scale labeled segmentation data. We therefore pretrain a universal segmenter that can be reused across tasks without segmentation supervision during adaptation.  
To achieve this, we annotate 8M videos~\cite{chen2024panda} and 15M images~\cite{sharma2018conceptual,changpinyo2021conceptual} with panoptic object trajectory masks generated by the TrajViT pipeline, which serve as pseudo ground truth.  
Optimization details are described in the supplementary material. 
\textit{This trained segmenter is reused in both \adaptername{} and \vlmname{}.}

\begin{table*}[h]
\centering
\setlength{\tabcolsep}{8pt} %

\begin{minipage}{0.63\textwidth}
    \centering
    \small
    \setlength{\tabcolsep}{3pt}
    \resizebox{\linewidth}{!}{%
    \begin{tabular}{lcccccccccccc}
        \toprule
        \multirow{2}{*}{\textbf{Model}} & 
        \multicolumn{2}{c}{\textbf{ActivityNet}~\cite{caba2015activitynet}} & 
        \multicolumn{2}{c}{\textbf{VATEX}~\cite{wang2019vatex}} &
        \multicolumn{2}{c}{\textbf{MSR-VTT}~\cite{xu2016msrvtt}} &
        \multicolumn{2}{c}{\textbf{Charades}~\cite{sigurdsson2016charades}} &
        \multicolumn{2}{c}{\textbf{COCO}~\cite{lin2014coco}} &
        \multicolumn{2}{c}{\textbf{Flickr30K}~\cite{plummer2015flickr30k}} \\
        \cmidrule(lr){2-3} \cmidrule(lr){4-5} \cmidrule(lr){6-7} \cmidrule(lr){8-9} \cmidrule(lr){10-11} \cmidrule(lr){12-13}
        & {\small txt2vid} & {\small vid2txt} 
        & {\small txt2vid} & {\small vid2txt} 
        & {\small txt2vid} & {\small vid2txt} 
        & {\small txt2vid} & {\small vid2txt}
        & {\small txt2img} & {\small img2txt}
        & {\small txt2img} & {\small img2txt} \\
        \midrule
        ViT3D        & 37.1 & 35.6 & 36.4 & 60.2 & 31.4 & 58.1 & 13.8 & 12.9 & 61.5 & 67.4 & 75.2 & 83.6 \\
        TokenLearner & 36.4 & 36.2 & 34.3 & 58.8 & 30.2 & 57.6 & 12.0 & 12.5 & 59.3 & 65.8 & 73.0 & 81.0 \\
        ViViT        & 33.9 & 33.2 & 34.0 & 57.6 & 29.6 & 55.2 & 12.3 & 11.8 & 57.8 & 64.1 & 71.6 & 79.3 \\
        RLT          & 35.9 & 35.0 & 35.0 & 58.4 & 30.5 & 57.1 & 12.2 & 12.3 & 59.6 & 66.1 & 72.8 & 81.2 \\
        \midrule
        \textbf{TrajViT}      & 38.4 & 38.1 & 36.0 & 61.1 & 31.6 & 61.2 & 14.7 & 14.6 & 63.2 & 69.0 & 77.4 & 85.1 \\
        \textbf{TrajViT-2} & \textbf{40.1} & \textbf{42.2} & \textbf{37.9} & \textbf{65.0} & \textbf{33.2} & \textbf{65.3} & \textbf{15.9} & \textbf{16.5} & \textbf{68.1} & \textbf{75.6} & \textbf{83.5} & \textbf{90.4} \\
        \bottomrule
    \end{tabular}}
    \vspace{-2mm}
    \captionof{table}{\textbf{Zero-shot video \& image retrieval performance (R@5).}  \encodername{}  consistently surpasses all baselines by a large margin.
    }
    \label{tab:retrieval}
\end{minipage}
\hfill
\begin{minipage}{0.33\textwidth}
    \centering
    \small
    \setlength{\tabcolsep}{4pt}
    \resizebox{\linewidth}{!}{%
    \begin{tabular}{lccccc} 
        \toprule
        \textbf{Model} & \textbf{K400}~\cite{kay2017kinetics} & \textbf{SSV2}~\cite{goyal2017ssv2} & \textbf{IN-1K}~\cite{deng2009imagenet} & \textbf{CIFAR-100}~\cite{krizhevsky2009learning} & \textbf{Caltech-101}~\cite{fei2004caltech101} \\
        \midrule
        ViT3D        & 54.2 & 46.3 & \textbf{69.4} & 52.1 & 88.2 \\
        TokenLearner & 52.9 & 42.4 & 67.3 & 49.5 & 86.7 \\
        ViViT        & 51.2 & 43.1 & 66.8 & 48.2 & 86.3 \\
        RLT          & 52.5 & 43.6 & 67.0 & 49.0 & 86.5 \\
        \midrule
        \textbf{TrajViT}      & 55.3 & 45.7 & 68.1 & 50.8 & 87.4 \\
        \textbf{TrajViT-2} & \textbf{59.1} & \textbf{48.7} & 67.2 & \textbf{52.6} & \textbf{88.5} \\
        \bottomrule
    \end{tabular}}
    \vspace{-2mm}
    \captionof{table}{\textbf{Action \& image classification (attentive probe, top-1)}. IN-1K stands for ImageNet-1K.\encodername{} improves upon TrajViT and outperforms all baselines in most of the benchmarks.}
    \label{tab:classification}
\end{minipage}

\label{tab:combined}
\end{table*}

\begin{figure*}[htbp]
    \centering
    \begin{minipage}{0.71\linewidth} %
        \centering
        \small %
        \includegraphics[width=\linewidth]{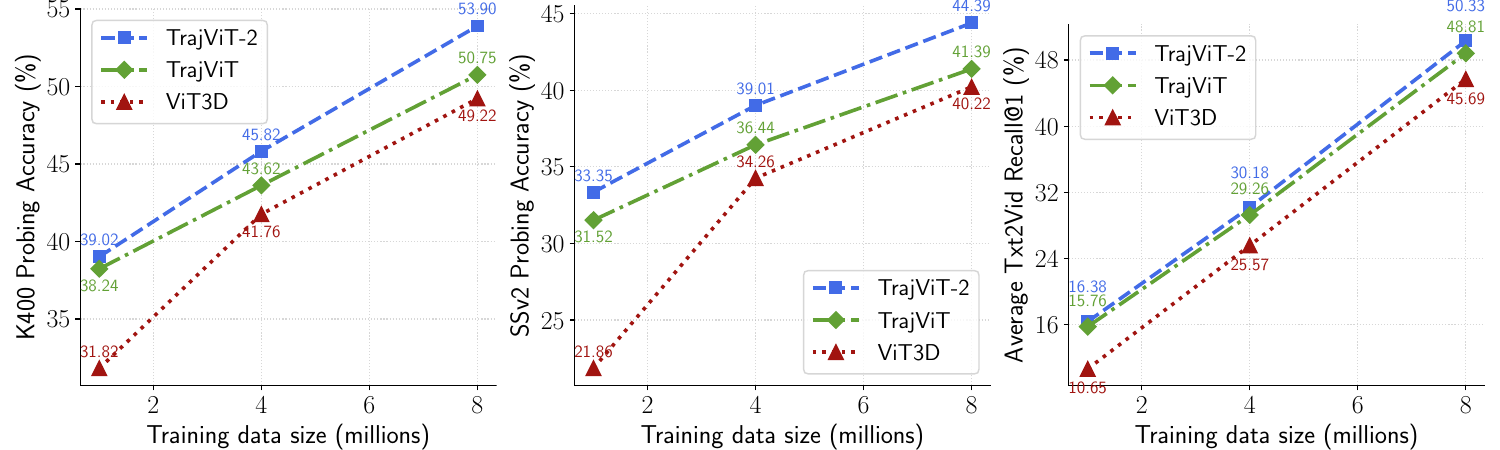}
            \vspace{-4mm}
      \caption{\textbf{Scaling with video training data.} \encodername{} exhibits stronger scaling behavior than TrajViT and sustains a consistent performance margin over ViT3D at every data scale.}
      \label{fig:datascale}
    \end{minipage}
    \hfill
    \begin{minipage}{0.275\linewidth} %
        \centering
        \small %
        \includegraphics[width=\linewidth]{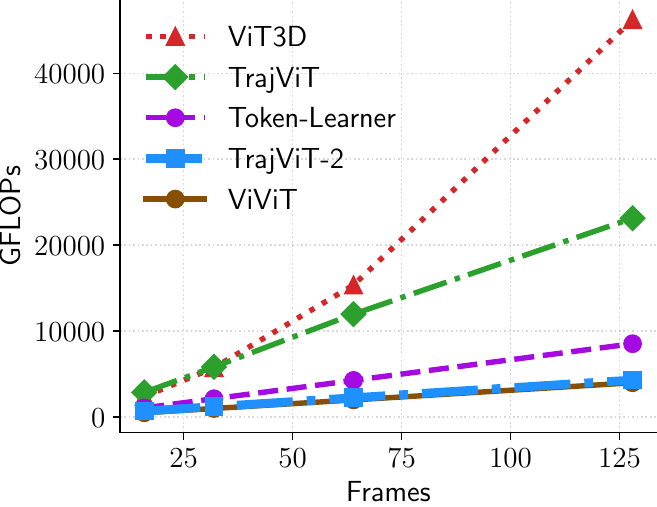}
            \vspace{-4mm}
        \caption{\textbf{Test time FLOPs comparison under different frame numbers.}}
        
        \label{fig:flops}
    \end{minipage}
\end{figure*}

\subsection{\encodername{}: A new video encoder}

We first consider the scenario where the tokenizer operates directly on raw visual inputs and the resulting trajectory tokens serve as input tokens for a transformer video encoder—identical to the setup used in TrajViT. We named the trained encoder \textbf{\encodername{}}.  
Following the same protocol, we jointly train our tokenizer and a large transformer encoder from scratch under the CLIP objective on a large-scale captioning corpus.  
Unlike TrajViT, which relies on an external pipeline to generate trajectories, our model learns them end-to-end by training the segmenter simultaneously with the transformer, with an additional segmentation supervision.

\noindent\textbf{Baselines.}
We compare \encodername{} with several representative architectures:  
(1) \textbf{ViT3D}, a standard video vision transformer that tokenizes inputs into fixed $16{\times}16{\times}2$ space–time patches;  
(2) \textbf{ViViT}~\cite{vivit}, a factorized video transformer that decouples spatial and temporal attention for efficient video modeling;  
(3) \textbf{TokenLearner}~\cite{tokenlearner}, which dynamically learns a compact set of informative tokens via learned attention pooling; and  
(4) \textbf{Run Length Tokenization (RLT)}~\cite{Choudhury2024RLT}, a token-merging approach that aggregates redundant patches based on similarity of patch pixels.  
All baseline models follow their original token number settings.

\noindent\textbf{Training and evaluation setup.}
 We train all models with visual-text contrastive learning objective (CLIP loss) from scratch. All models adopt the same size transformer as in ViT-Large.
Training corpus contains 4M video clips randomly sampled from Panda-70M~\cite{chen2024panda} and 15M image–caption pairs from CC3M~\cite{changpinyo2021conceptual} and CC12M~\cite{sharma2018conceptual}.
During training, we uniformly sample 8 frames per video, while in evaluation we uniformly sample 16 frames.
We use a global batch size of 1024 images and 128 videos for 20 epochs on 8 A100 GPUs. 
After pretraining, all encoders are frozen and evaluated on a broad set of visual understanding benchmarks spanning both video and image domains:   
Video-text retrieval is evaluated on ActivityNet~\cite{caba2015activitynet}, VATEX~\cite{wang2019vatex}, MSR-VTT~\cite{xu2016msrvtt}, and Charades~\cite{sigurdsson2016charades}; image-text retrieval is measured on COCO~\cite{lin2014coco} and Flickr30K~\cite{plummer2015flickr30k};  
For image and video classification, we perform linear probing on Kinetics-400 (K400)~\cite{kay2017kinetics}, Something-Something V2 (SSV2)~\cite{goyal2017ssv2}, ImageNet-1K (IN-1K)~\cite{deng2009imagenet}, CIFAR-100~\cite{krizhevsky2009learning}, and Caltech-101~\cite{fei2004caltech101}.

\noindent\textbf{\encodername{} performs better than all baselines.}
As shown in Table~\ref{tab:retrieval} and Table~\ref{tab:classification}, \encodername{} achieves consistent improvements over TrajViT and outperforms all baselines across both retrieval and classification benchmarks.
Compared with TrajViT, it attains higher recall on all retrieval datasets (e.g., +4.1\% vid2txt R@5 on ActivityNet and +4.0\% on VATEX) and stronger accuracy on video and image classification tasks (e.g., +3.8\% on K400 and +3.0\% on SSV2).  
On ImageNet, however, \encodername{} performs slightly lower than ViT3D.  
This is likely because ImageNet images typically contain a single centered foreground object and simple background, causing the segmenter to produce too few segments and therefore fewer tokens, which limits fine-grained discrimination on such easy scenes.  
Despite this, \encodername{} matches or surpasses all other baselines on cross-domain and multi-object datasets, underscoring the strength of its trajectory-level tokenization.

\noindent\textbf{\encodername{} scales better.}  
A key limitation of TrajViT lies in its scalability: its performance gain over ViT3D diminishes substantially as the pretraining dataset size increases from 1M to 8M samples.  
To examine the data-scaling behavior of \encodername{}, we follow the same experimental protocol used in TrajViT by partitioning the Panda-10M dataset into three random subsets containing 1M, 4M, and 8M video clips.  
We train \encodername{}, TrajViT, and ViT3D on all three scales and report their performance on video benchmarks.  
As shown in Figure~\ref{fig:datascale}, \encodername{} exhibits a much stronger scaling trend than TrajViT.
At the largest scale, \encodername{} continues to outperform ViT3D by a large margin across both classification and retrieval tasks.
We attribute this improvement to the end-to-end differentiability of \encodername{}’s tokenizer: the segmenter can flexibly adjust its segmentation behavior in response to the pretraining objective, rather than relying on fixed, heuristic segmentations. We give qualititative illustration in Figure~\ref{fig:seg_clip}.

\noindent\textbf{\encodername{} is highly efficient.}
Another drawback of TrajViT is its heavy computational overhead, caused by dependence on an external pipeline.  
\encodername{} resolves this issue by replacing it with a lightweight, fully integrated segmenter.  
Our entire trajectory tokenizer contains only 46M parameters—an order of magnitude smaller compared to the 304M parameters of the ViT-Large backbone.  
We further compare inference FLOPs across input frame counts from 16 to 128 in Figure~\ref{fig:flops}.
\encodername{} achieves nearly the same computational cost as the most efficient baseline, ViViT, in stark contrast to the quadratic scaling of patch-based ViT3D and the high-slope linear scaling of TrajViT.  
These results demonstrate that \encodername{} achieves superior efficiency while maintaining strong performance.

\begin{table}[t]
\centering
\setlength{\tabcolsep}{8pt}
\resizebox{\linewidth}{!}{%
\begin{tabular}{lcccc}
\toprule
& \multicolumn{2}{c}{\textbf{VideoMAE-v2~\cite{fu2024videomme}}} & \multicolumn{2}{c}{\textbf{V-JEPA2~\cite{assran2025v}}} \\
\cmidrule(lr){2-3} \cmidrule(lr){4-5}
\textbf{Probing Method} & \textbf{K400~\cite{kay2017kinetics}} & \textbf{SSv2~\cite{goyal2017ssv2}} & \textbf{K400~\cite{kay2017kinetics}} & \textbf{SSv2~\cite{goyal2017ssv2}} \\
\midrule
Linear probing               & 79.4 & 59.1 & 84.5 & 73.7 \\
Attentive probing            & 80.2 & 59.7 & 85.1 & 74.2 \\
Perceiver probing            & 79.9 & 59.8 & 84.7 & 74.2 \\
\adaptername{} (1 token/traj)   & 82.0 & 60.4 & 87.2 & 74.6 \\
\adaptername{} (2 token/traj)   & 82.4 & 60.8 & 87.8 & 74.9 \\
\adaptername{} (4 token/traj)   & \textbf{82.5} & \textbf{60.9} & \textbf{88.0} & \textbf{75.1} \\
\bottomrule
\end{tabular}
}
\caption{\textbf{Probing results on different video backbones.}  
Top-1 accuracy (\%) on Kinetics-400 (K400) and Something-Something-V2 (SSv2) using various probing strategies.  
\adaptername{} consistently improves the downstream probing accuracy of pretrained backbone features, and performance further increases with more tokens per trajectory.}
\vspace{-15pt}
\label{tab:probing}
\end{table}

\begin{figure*}[htbp]
    \centering
    \includegraphics[width=\linewidth]{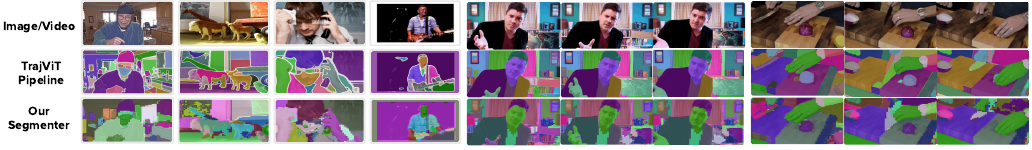}
     \caption{\textbf{Trajectory masks produced by our segmenter vs.  by TrajViT pipeline.} While our segmenter produces coarser masks and may miss very small objects, it demonstrates strong semantic grouping ability that is sufficient for downstream understanding tasks. 
}
    \label{fig:gt_seg}
\end{figure*}

\subsection{\adaptername{}: A new video probing head}

In practice, as pretrained large vision encoders continue to improve, it becomes increasingly desirable to directly reuse their output feature maps—often dense patch-level tokens—for downstream tasks.  
We show that \modulename{} can be directly plugged in as a lightweight adapter module to reorganize these dense feature maps into a compact set of trajectory tokens.  
We show this design not only reduces token number for downstream models but also provides a cost-effective way to enhance the probing performance of in downstream tasks without full fine-tuning.

\noindent\textbf{Training and evaluation setup.}  
We take video action classification as an example to demonstrate this setting.  
As illustrated in the second part of Figure~\ref{fig:Applications}, \modulename{} is inserted after a frozen ViT backbone to reorganize output tokens, which are then used by an attentive probing head to predict classification logits.
The segmenter is pretrained and \textit{kept frozen} during probing, while the trajectory encoder is trained jointly with the probing head.  
For pretrained backbones, we adopt VideoMAE-v2~\cite{wang2023videomae} and V-JEPA-2~\cite{assran2025v}, both using their ViT-Huge variants.  
We evaluate action recognition accuracy on the Kinetics-400 and Something-Something V2 (SSv2) benchmarks.
Videos are uniformly sampled to 16 frames and sent to segmenter in one forward pass, producing a maximum of 128 trajectory tokens.

We compare our approach against three baselines: (1) naive linear probing, (2) attentive probing without adaptation, and (3) a Perceiver module of identical size and number of learnable queries as our trajectory encoder but without trajectory priors.  
In addition, we enable the adaptive token number mechanism in the trajectory encoder and report results for varying numbers of tokens per trajectory.

\noindent\textbf{Results.}  
Table~\ref{tab:probing}
summarizes the top-1 classification accuracy of different probing strategies.  
Compared to both linear and attentive probing, \adaptername{} consistently achieves higher accuracy across datasets. 
Furthermore, our method outperforms the Perceiver-only variant, indicating that the improvement arises not merely from additional parameters but from the incorporation of trajectory priors.  
In fact, naively inserting a Perceiver module does not yield any gain over the attentive probing baseline.
We also observe a steady performance increase as the token number per trajectory grows, even though the single-token configuration already surpasses conventional probing methods.  
These results demonstrate that the proposed tokenizer is not only effective for end-to-end video representation learning, but also serves as a plug-in adapter that enhances features in pretrained ViT backbones in a parameter-efficient manner.

\vspace{0.5em}

\begin{figure*}[t]
    \centering
    \includegraphics[width=\linewidth]{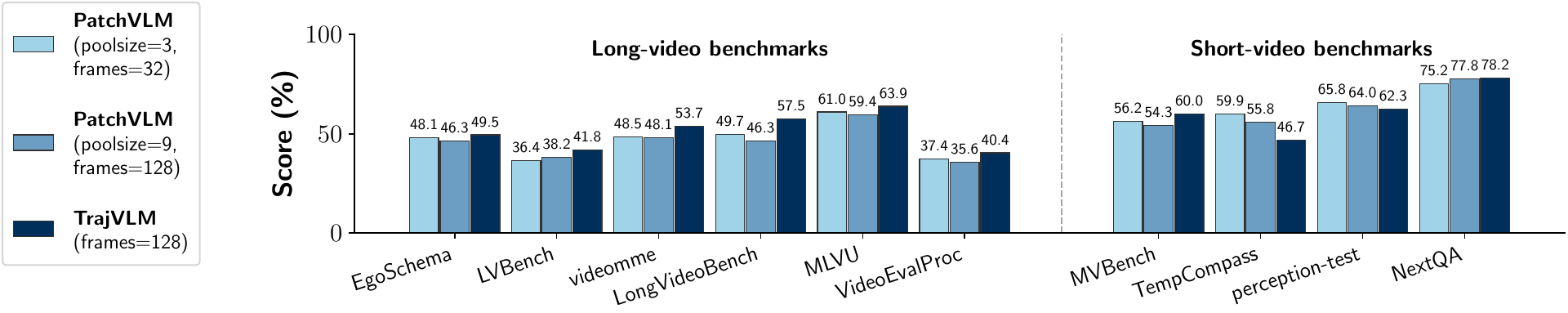}
    \vspace{-20pt}
    \caption{\textbf{VideoQA results for \modulename{} applying to large vision-language model.} VLM with \modulename{} as connector (\vlmname{}) notably outperforms patch pooling baseline (PatchVLM) in long-video benchmarks, while the performance is mixed for short-video benchmark.}
    \label{fig:vlm_result}
    \vspace{-10pt}
\end{figure*}

\subsection{\vlmname{}: A new video-language model}

Finally, we demonstrate that ~\modulename{} can also serve as a connector between a vision encoder and a language model, providing an object-centric alternative to the patch-pooling connectors commonly used in large vision–language models (VLMs)~\cite{liu2023visual,Qwen2024Qwen25VL,prunevid,Deitke2024Molmo}.  
To this end, we build a small-scale model named \textbf{\vlmname{}} by integrating our tokenizer into a standard LLaVA-style architecture (Figure~\ref{fig:Applications} part 3).
The goal of this experiment is not to compete with state-of-the-art VLMs, but rather to provide an \textit{apple-to-apple} comparison between two connector designs: TrajTok and commonly-used patch pooling.  
Scaling \vlmname{} to larger models with increased compute remains a future direction.

\noindent\textbf{Architecture and Baseline.}  
We use \textbf{Qwen3-4B}~\cite{yang2025qwen3} as the language model backbone and \textbf{SigLIP2-Huge}~\cite{tschannen2025siglip} as the vision encoder.  
For the baseline connector, we follow the design of  \textbf{Molmo}~\cite{Deitke2024Molmo}, which employs per-frame, patch-based attention pooling.  
Specifically, each $x \!\times\! x$ spatial patch window is pooled into a single vector via a multi-head attention layer, where the mean patch embedding serves as the query.  
Notably, similar patch-pooling strategies are widely used in many of today’s most widely-used open-source vision–language models~\cite{Qwen2024Qwen25VL,xu2024slowfast,wang2025internvl3}.

\noindent\textbf{Training Data and Recipe.}  
We adopt a subset of Molmo-2’s training corpus~\cite{clark2025molmo2}, including the PixMo captioning split, synthetic VideoQA split, and academic QA datasets (details in supplementary).  
Training follows Molmo’s two-stage procedure:
\begin{itemize}
    \item \textbf{Pretraining.} All parameters are pretrained on the PixMo captioning split for one epoch to align  visual features with the language model.
    \item \textbf{Fine-tuning.} The model is then fine-tuned for 10{,}000 steps on the remaining QA datasets.  
\end{itemize}
All experiments are conducted on 8~A100 GPUs with a sequence length of 8{,}192 tokens. For \vlmname{} frame sampling, we uniformly sample 128 video frames during both training and evaluation.  
\modulename{} connector processes 128 frames by truncating them into 16-frame clips, each proposing a maximum of 128 tokens. 
For baseline PatchVLM, we train two versions of the model: a version that uses common patch pooling size $x=3$, but can only support 32 frames due to sequence length limits; Another version that uses patch pool size $x=9$ so that it can support 128-frame during training and the resulting number of visual tokens roughly matches that of our trajectory tokenizer, ensuring a fair comparison.

\noindent\textbf{Results.}  We evaluate VLMs in common video QA benchmarks~\cite{ma2025videoevalpro,liu2024tempcompass,patraucean2023perceptiontest,xiao2021nextqa,li2024mvbench,mangalam2023egoschema,wu2024longvideobench,fu2024videomme,zhang2024longvideobench,zhou2025mlvu}. Figure~\ref{fig:vlm_result} shows that \vlmname{} consistently outperforms the patch-pooling baselines on long-video benchmarks, including a notable +8.8\% on LongVideoBench and +5.4\% on LVBench over PatchVLM with default poolsize=3. We attribute our improvement to the the fact that  \modulename{} produces semantically structured tokens that better support long-range reasoning while reducing redundancy. Notably, increasing the pooling window size in PatchVLM does not improve long-video performance on most benchmarks, indicating that naively trading off spatial resolution with temporal support is insufficient for long-range reasoning.
For Short-video benchmarks, we observe increased performance on \cite{xiao2021nextqa,li2024mvbench} bur decreased performance on \cite{liu2024tempcompass,patraucean2023perceptiontest}.  
Overall, these results validate \modulename{} as an effective connector for VLMs, particularly in long-video understanding.

\begin{table}[t]
\centering
\resizebox{0.99\linewidth}{!}{
{\def\arraystretch{1.12}
\begin{tabular}{ll|ccc}
\toprule
\textbf{Module} & \textbf{Variation} & \textbf{VEQ (\%)} & \textbf{STQ (\%)} & \textbf{Retrieval (R@5)} \\
\midrule
\multicolumn{2}{c|}{\textbf{Default Architecture}} 
& \baseline{42.3} & \baseline{70.1} & \baseline{22.1} \\
\midrule
Backbone & no hierarchical features
& 39.3 \textcolor{gray}{\footnotesize (\(\downarrow\) 3.0)} 
& 66.2 \textcolor{gray}{\footnotesize (\(\downarrow\) 3.9)} 
& 19.2 \textcolor{gray}{\footnotesize (\(\downarrow\) 2.9)} \\
Output Res. & 56\(\rightarrow\)224
& 44.1 \textcolor{gray}{\footnotesize (\(\uparrow\) 1.8)} 
& 73.0 \textcolor{gray}{\footnotesize (\(\uparrow\) 2.9)} 
& 22.0 \textcolor{gray}{\footnotesize (\(\downarrow\) 0.1)} \\
Perceiver & no detach gradient
& 34.1 \textcolor{gray}{\footnotesize (\(\downarrow\) 8.2)} 
& 59.3 \textcolor{gray}{\footnotesize (\(\downarrow\) 10.8)} 
& 18.3 \textcolor{gray}{\footnotesize (\(\downarrow\) 3.8)} \\

\midrule
Seg. Loss & \textminus{} dice loss 
& 39.0 \textcolor{gray}{\footnotesize (\(\downarrow\) 3.3)} 
& 68.9 \textcolor{gray}{\footnotesize (\(\downarrow\) 1.2)} 
& 16.7 \textcolor{gray}{\footnotesize (\(\downarrow\) 5.4)} \\
Seg. Loss & \textminus{} focal loss
& 41.2 \textcolor{gray}{\footnotesize (\(\downarrow\) 1.1)} 
& 67.4 \textcolor{gray}{\footnotesize (\(\downarrow\) 2.7)} 
& 22.1 \textcolor{gray}{\footnotesize (0.0)} \\
Seg. Loss & {+} cross-entropy loss
& 42.1 \textcolor{gray}{\footnotesize (\(\downarrow\) 0.2)} 
& 71.3 \textcolor{gray}{\footnotesize (\(\uparrow\) 1.2)} 
& 21.3 \textcolor{gray}{\footnotesize (\(\downarrow\) 0.8)} \\
\bottomrule
\end{tabular}}}
\vspace{-1mm}
\caption{\textbf{Ablations of segmenter design.}}
\label{tab:abl_segmenter}
\vspace{-2mm}
\end{table}

\begin{table}[t]
\centering
\resizebox{0.99\linewidth}{!}{
{\def\arraystretch{1.12}
\begin{tabular}{ll|ccc}
\toprule
\textbf{Module} & \textbf{Variation} & \textbf{1 token/traj} & \textbf{2 token/traj} & \textbf{4 token/traj} \\
\midrule
\multicolumn{2}{@{}c@{}}{\textbf{Default Architecture}} 
& \baseline{22.1} & \baseline{23.0} & \baseline{23.2} \\
\midrule
Attention Mask & w/o mask 
& 17.4 \textcolor{gray}{\footnotesize (\(\downarrow\) 4.7)} 
& 17.9 \textcolor{gray}{\footnotesize (\(\downarrow\) 5.1)} 
& 18.3 \textcolor{gray}{\footnotesize (\(\downarrow\) 4.9)} \\
Query Init. & Fourier \(\rightarrow\) random 
& 22.1 \textcolor{gray}{\footnotesize (0.0)} 
& 22.5 \textcolor{gray}{\footnotesize (\(\downarrow\) 0.5)} 
& 21.9 \textcolor{gray}{\footnotesize (\(\downarrow\) 1.3)} \\
Perceiver Depth & 2 \(\rightarrow\) 4 
& 22.3 \textcolor{gray}{\footnotesize (\(\uparrow\) 0.2)} 
& 22.9 \textcolor{gray}{\footnotesize (\(\downarrow\) 0.1)} 
& 23.5 \textcolor{gray}{\footnotesize (\(\uparrow\) 0.3)} \\
\bottomrule
\end{tabular}}}
\vspace{-1mm}
\caption{\textbf{Ablations of trajectory encoder design.} }
\label{tab:abl_encoder}
\vspace{-5mm}
\end{table}

\section{Ablating \modulename{} design}
\label{sec:abltion}
We ablate the design choices of \modulename{} under \encodername{} setting, which trains a video encoder jointly optimized by the segmentation loss and CLIP loss.  
All experiments are trained on 1M video–caption pairs randomly sampled from Panda-10M for 10 epochs using 4 GPUs.

\noindent\textbf{Ablation of segmenter design.} 
We ablate the major design choices of the segmenter, including backbone hierarchy, gradient detachment, output resolution, and segmentation loss functions.  We use VEQ and STQ metrics to quantify video panoptic segmentation quality following~\cite{ye2025entitysam}, and use average txt2vid R@5 accuracy across video retrieval benchmarks~\cite{caba2015activitynet,wang2019vatex,xu2016msrvtt,sigurdsson2016charades} to quantify video understanding performance.
As summarized in Table~\ref{tab:abl_segmenter}, removing hierarchical features yields consistent drops across VEQ, STQ, and retrieval accuracy, confirming the importance of multi-scale representations.  
Removing the gradient detachment of the patch-feature inside the perceiver causes large declines in segmentation quality due to coupled updates between queries and patch features.  
Increasing the output resolution slightly improves VEQ/STQ,  but has negligible impact on retrieval, indicating that coarse masks are sufficient for semantic grouping.  
Among loss components, dice loss is the most critical: removing it severely harms both segmentation and understanding performance.

\noindent\textbf{Ablation of trajectory encoder design.} 
We also ablate the key components of the trajectory encoder, including the use of attention masks, query initialization strategies, and perceiver depth under the same retrieval task (Table~\ref{tab:abl_encoder}).  
Removing the hard attention mask significantly degrades performance by weakening the association between trajectory tokens and their assigned regions.  
For query initializatoin under multi-token per trajectory setting,  increasing the number of tokens per trajectory would not improve performance if Fourier-based query initialization is replaced by random initialization. That might be because different slots might extract the same trajectory information without explicit diversity encouragement. 
Finally, increasing perceiver depth yields only marginal improvements at higher computational cost, suggesting that a shallow perceiver is already sufficient for capturing local dynamics within trajectories.

\section{Conclusion}

We introduces \modulename{}, an end-to-end and efficient tokenizer that learns to group visual trajectories and produce trajectory-level tokens directly from video inputs.
Our experiments show that \modulename{} is high-performance and highly versatile—it improves performance in three scenarios of
pretraining video encoder, probing pretrained features, and training a video-language model.
These results highlight the potential of trajectory-based tokenization as a more efficient and semantically aligned alternative to traditional patchification.

\noindent\textbf{Acknowledgement.} This project was funded by DSO National Laboratories in Singapore and by Toyota Motor Inc.

\maketitlesupplementary

\section{Segementer Training Details}

In the \textbf{TrajAdapter} and \textbf{TrajVLM} settings, we pretrain the trajectory segmenter once and reuse its weights for initialization during downstream probing and VLM training.  
This section provides full details of the dataset construction, annotation pipeline, filtering criteria, and training configuration for our segmenter training..

\subsection{Dataset Construction}
\label{sec:seg_training}
\noindent \textbf{Sources.}
We construct a video \& image corpus for segmenter training by combining:
Panda (video)~\cite{chen2024panda}, CC12M (image)~\cite{changpinyo2021conceptual}, CC3M (image)~\cite{sharma2018conceptual}, and a subset of DataComp-50M (image)~\cite{gadre2023datacomp}.  
All samples are annotated with pseudo panoptic trajectory masks using the TrajViT trajectory–generation pipeline, followed with data filtering. We describe the details below.

\noindent \textbf{Annotation Pipeline.}
We adopt the same annotation process as in the TrajViT paper~\cite{zheng2025trajvit}.  
In summary, the pipeline consists of four steps.
\begin{enumerate}
    \item sample frames and detect keyframes based on feature changes in colorspace and Luminance Histogram. 
    \item generate panoptic object masks in the key frames using DirectSAM~\cite{chen2024subobject} model.
    \item track objects across frames via SAM2~\cite{ravi2024sam2}. 
    \item merge instance masks between CLIPs using heuristics like IOU overlaps to form long-term trajectories.  
\end{enumerate}
The pipeline uses external models like DirectSAM and SAM2~\cite{chen2024subobject,ravi2024sam2}.  
For images, only spatial segmentation steps are applied.

\noindent \textbf{Quality Filtering.}
We apply two filtering criteria to remove low-quality pseudo labels:
\begin{itemize}
    \item \textbf{Coverage filter:} remove samples where the union of all trajectory masks covers less than 80\% of pixels.
    \item \textbf{Object-count filter:} remove samples containing fewer than 10 detected objects.
\end{itemize}
After filtering, we retain roughly 2.5M images and 2.0M videos for segmenter pretraining.

\subsection{Training Configuration}

The segmenter is trained on the filtered dataset.  Different from \encodername{} where all modules are trained from scratch, we initialize the ConvNext-small patch encoder from DINOv3's weights, which helps in the generalization performance of produced segments. Other modules are initialized from scratch.
We train the model for 20 epochs with  8 A100 GPUs. We use the base learning rate of $1{\times}10^{-3}$, and adopt a linear decay learning-rate schedule with warm-up.   Additional hyperparameters are summarized in Table~\ref{tab:segmenter_hyperparams}.

\begin{table}[h]
\centering
\setlength{\tabcolsep}{8pt}
\resizebox{0.95\linewidth}{!}{
\begin{tabular}{l|c}
\toprule 
\textbf{Hyperparameter} & \textbf{Value} \\
\midrule
Resolution & 224 \\
Frame sampling & uniform 16 frames (videos only) \\
Optimizer & AdamW \\
Base LR & $1\times10^{-3}$ \\
Weight decay & 0.01 \\
Optimizer momentum & $\beta_1=0.9,\ \beta_2=0.999$ \\
Batch size & video-128,\ image-512 \\
Training epochs & 20 \\
LR schedule & linear decay \\
Warm up epochs & 1 \\
Warm up schedule & linear warm-up \\
Random crop scale & (0.2,\ 1.0) \\
Random crop ratio & (3/4,\ 4/3) \\
Horizontal flip probability & 0.5 \\
Color jitter probability & 0.8 \\
Gaussian blur probability & 0.5 \\
Grayscale probability & 0.2 \\
\bottomrule
\end{tabular}
}
\caption{\textbf{Hyperparameters used for segmenter pretraining.}  }
\label{tab:segmenter_hyperparams}
\end{table}

\section{More Qualitative Examples of the Segmenter}
\begin{figure*}
    \centering
    \includegraphics[width=\linewidth]{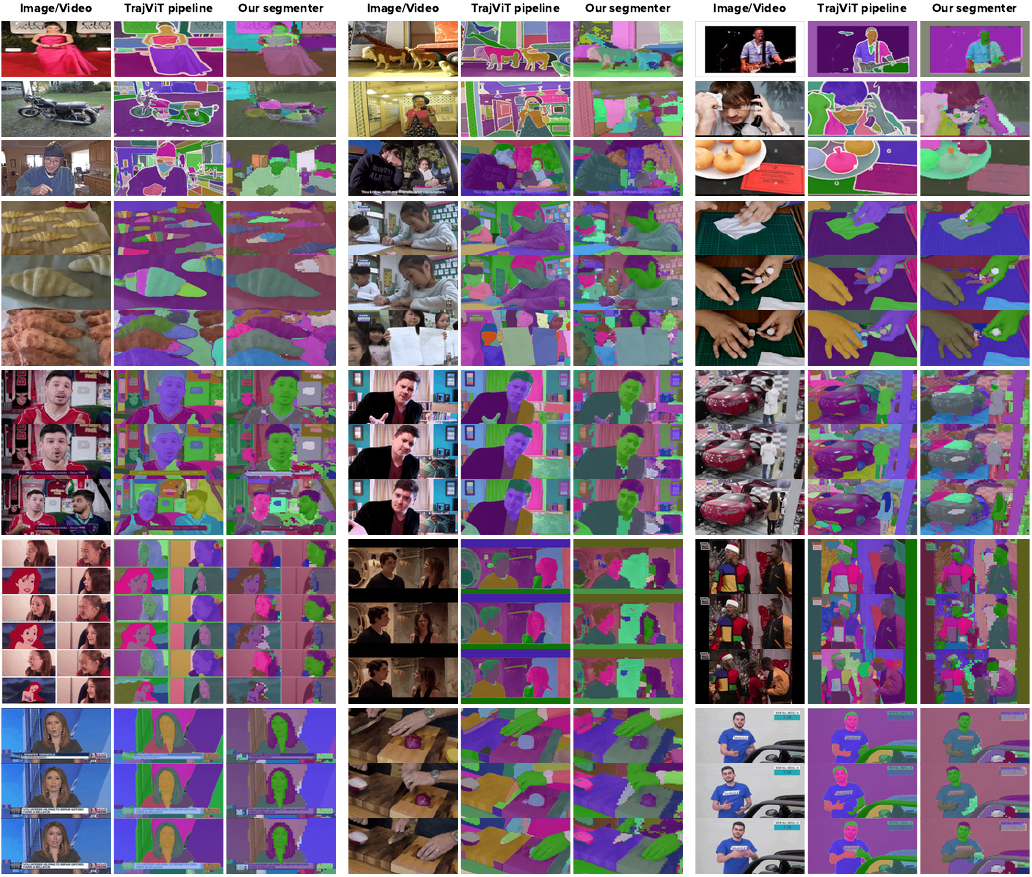}
    \caption{Qualitative Examples of the trajectory masks produced by our segmenter.}
    \label{fig:all_examples}
\end{figure*}

We show examples of generated trajectories from our training set in Fig.~\ref{fig:all_examples}. Overall, the segmenter exhibits strong semantic grouping ability, consistently discovering object-level regions that are sufficiently accurate for downstream understanding tasks.
From the perspective of pixel-level segmentation quality, however, the lightweight design and low output resolution introduce several expected limitations: the model occasionally misses very small objects, may over-merge background regions, and produces imprecise object boundaries. These imperfections, while noticeable visually, do not hinder its effectiveness as a trajectory proposal module, as our downstream tasks primarily rely on correct semantic grouping rather than pixel-perfect masks.

\section{Quantitative Evaluation of the Segmenter}

Although the proposed segmenter in the main paper is intentionally lightweight—prioritizing semantic grouping over pixel-level precision—we additionally study how well it can perform on the standard panoptic video segmentation task when its capacity is scaled up.  
This experiment is conducted purely for analysis and is \emph{not} used by any model in the main paper.

\noindent \textbf{Scaling up the segmenter.}
We keep the same training dataset as described in Sec.~\ref{sec:seg_training}, but increase the segmenter capacity in two ways:
(1) replacing the ConvNeXt-Tiny patch encoder with a ConvNeXt-Large backbone and expanding the Perceiver stack from 2 layers to 4 layers, and  
(2) producing full-resolution predictions by adding a pixel decoder identical to the one used in SAM~\cite{ravi2024sam2}, applied on top of the downsampled patch features.  
The input and output resolution are both set to $512\times512$.

\noindent \textbf{Benchmark and competitors.}
We evaluate on the ViPEntitySeg~\cite{miao2022large} benchmark and compare against two state-of-the-art video panoptic segmentation systems: EntitySAM~\cite{ye2025entitysam} and SAM~2.1~\cite{ravi2024sam2}.  
We report VEQ-SQ, VEQ-RQ, and STQ-EN following the benchmark protocol.

\noindent \textbf{Results.}
Table~\ref{tab:vipseg_eval} shows that the scaled-up version of our segmenter achieves competitive performance, surpassing EntitySAM in VEQ-SQ and improving VEQ-RQ relative to SAM~2.1.  
While its STQ-EN score is slightly lower than EntitySAM, these results demonstrate that our grouping-centric design can approach state-of-the-art performance when augmented with a strong visual backbone and a full-resolution decoder, confirming our segmenter design is reasonable.

\begin{table}[h]
\centering
\setlength{\tabcolsep}{8pt}
\resizebox{0.85\linewidth}{!}{
\begin{tabular}{lccc}
\toprule
\textbf{VIPSeg Benchmark} & \textbf{VEQ-SQ} & \textbf{VEQ-RQ} & \textbf{STQ-EN} \\
\midrule
EntitySAM & 84.7 & \textbf{64.5} & \textbf{43.3} \\
SAM~2.1    & 83.1 & 36.7 & 41.7 \\
Ours (scaled-up) & \textbf{85.5} & 45.1 & 40.2 \\
\bottomrule
\end{tabular}
}
\caption{\textbf{Evaluation of a scaled-up version of our segmenter on the ViPEntitySeg benchmark.}  
This model is \emph{not} used in any main-paper experiments; it serves only as an isolated study of segmentation quality under increased capacity.}
\label{tab:vipseg_eval}
\end{table}

\section{Training Details for \encodername{}}

For all \encodername{} experiments and baseline models, we optimize using the AdamW optimizer~\cite{loshchilov2019adamw} with a base learning rate of $10^{-4}$, weight decay of $10^{-2}$, and mixed-precision training. We use a cosine annealing schedule with a linear warm-up of one epoch. The contrastive batch size is 128 for video clips and 1024 for images. All models are trained for 20 epochs using 8 NVIDIA A100 GPUs.
During training, we apply standard video augmentation including random ColorJitter, Grayscale, Gaussian blur, horizontal flip, and resized cropping. At evaluation, we use only a single resizing operation for consistency. All models adopt a ViT-Large transformer and operate on 224-resolution inputs with 16 uniformly sampled frames.

\section{Training Details for \adaptername{}}

For all \adaptername{} experiments, we follow the standard protocol for probing pretrained video encoders. We use the AdamW optimizer with a learning rate of $1\times 10^{-4}$ and weight decay of $0.5$. The pretrained backbone is kept frozen, while the trajectory encoder and probing head are updated. Before classification, video features are layer-normalized. 
We train with a batch size of 128 for 10 epochs. This configuration is used for all \adaptername{} experiments on both Kinetics-400 and Something-Something-V2 probing tasks.

\section{Training Details for \vlmname{}}
We provide more training details for \vlmname{} in this section.

\noindent\textbf{Data sources.} As discussed in the main paper, \vlmname{} is trained using a two-stage procedure. 
For the pretraining stage, we closely follow the Molmo training paradigm~\cite{Deitke2024Molmo} and use the same PixMo captioning split to align visual representations with the language model.
For the instruction-tuning stage, we adopt the mixture of public academic VideoQA datasets and synthetic QA pairs curated in Molmo-2~\cite{clark2025molmo2}. 
These datasets span a wide range of reasoning skills—including temporal grounding, causal inference, long-horizon understanding, and multi-step procedural reasoning—and are summarized in Table~\ref{tab:vqa}. 
In total, the mixture contains approximately 5 million training examples.

\noindent\textbf{Training hyperparameters.} The training hyperparaemters of the first stage follows Molmo~\cite{Deitke2024Molmo}. We summarize the training hyperparameters for the second stage at Table~\ref{tab:trajvlm_hparams}.

\begin{table}[h!]
\centering
\caption{\textbf{Datasets used for training \vlmname{} under the ``final'' mixture.} 
This mixture combines a large set of academic VideoQA datasets, temporal reasoning datasets, and synthetic captioning/QA corpora.}
\vspace{0.5em}
\resizebox{\linewidth}{!}{
\begin{tabular}{l|l|c}
\toprule
\textbf{Category} & \textbf{Dataset Name(s)} & \textbf{Notes / Source} \\
\midrule

\multirow{11}{*}{\textbf{Academic VideoQA}}
& \texttt{llava\_video\_mc\_academic} & MC-style QA \\
& \texttt{llava\_video\_oe\_academic} & Open-ended QA \\
& \texttt{clevrer} & Causal \& counterfactual reasoning \\
& \texttt{funqa} & Fine-grained temporal QA \\
& \texttt{star} & Long-horizon procedural QA \\
& \texttt{intent\_qa} & Human intent reasoning \\
& \texttt{tgif} & Action/state transition QA \\
& \texttt{video\_localized\_narratives} & Localized narrations \\
& \texttt{road\_text\_vqa} & Driving VQA \\
& \texttt{countix\_oe} & Counting QA (open-ended) \\
& \texttt{camerabench\_qa} & Camera-motion VQA \\
\midrule

\multirow{12}{*}{\textbf{Action / Activity QA}}
& \texttt{nextqa\_mc} & Next-QA multiple-choice \\
& \texttt{news\_video\_qa\_filtered} & News comprehension QA \\
& \texttt{how2qa} & How-to instructional QA \\
& \texttt{sutd\_trafficqa} & Traffic event QA \\
& \texttt{social\_iq2} & Social reasoning \\
& \texttt{sportsqa\_oe} & Sports QA (OE) \\
& \texttt{cinepile} & Movie understanding QA \\
& \texttt{ssv2\_qa} & Something-Something QA \\
& \texttt{moments\_in\_time\_qa} & Activity recognition QA \\
& \texttt{kinetics\_qa} & Kinetics QA \\
& \texttt{charades\_sta\_all\_qa} & Charades Spatial-Temporal QA \\
& \texttt{coin\_all\_qa} & Procedural task step QA \\
\midrule

\multirow{6}{*}{\textbf{Video Captioning / Highlighting}}
& \texttt{youcook2\_all\_qa} & Recipe video QA/caption \\
& \texttt{activitynet\_all\_qa} & ActivityNet QA/caption \\
& \texttt{ego4d\_all} & Ego4D narrations + QA \\
& \texttt{video\_localized\_narratives\_caption} & Captioning corpus \\
& \texttt{qv\_highlights} & Highlight detection w/ text \\
& \texttt{motionbench\_train} & Long-range motion reasoning \\
\midrule

\multirow{3}{*}{\textbf{Internal / Synthetic VideoQA}}
& \texttt{vixmo\_syn\_video\_capqa\_v2} & 200K synthetic QA pairs \\
& \texttt{vixmo3\_top\_level\_captions\_min\_3} & 101K curated human captions \\
& \texttt{vixmo\_clip\_qa\_all} & CLIP-constructed QA corpus \\
\bottomrule
\end{tabular}
}

\label{tab:vqa}
\end{table}

\begin{table}[h]
\centering
\setlength{\tabcolsep}{8pt}
\resizebox{0.95\linewidth}{!}{
\begin{tabular}{l|c}
\toprule 
\textbf{Hyperparameter} & \textbf{Value} \\
\midrule
Max video frames & 128 \\
Total training steps & 10{,}000 \\
Training stages & Pretraining + Instruction tuning \\
Sequence length & 8192 \\
Global batch size & 32 \\
Device batch size & 4 \\
Number of GPUs & 8\,$\times$\,A100 (80GB) \\
Precision & bfloat16 (AMP) \\
Optimizer & AdamW \\
LLM learning rate & $1\times10^{-5}$ \\
ViT learning rate & $5\times10^{-6}$ \\
Connector learning rate & $5\times10^{-6}$ \\
Learning rate warmup & 200 steps \\
LR schedule & Multimodal cosine decay \\
Weight decay & 0.0 (LLM / ViT / connector) \\
Adam momentum & $\beta_1 = 0.9,\ \beta_2 = 0.95$ \\
Adam $\epsilon$ & $1\times10^{-6}$ \\
Gradient clipping & 1.0 \\
\bottomrule
\end{tabular}
}
\caption{\textbf{Key hyperparameters used for training \vlmname{}.} 
Values reflect the shared configuration across all VLM experiments.}
\label{tab:trajvlm_hparams}
\end{table}

{
    \small
    \bibliographystyle{ieeenat_fullname}
    \bibliography{main}
}
\end{document}